# Machine Learning-Based Security Policy Analysis

Krish Jain
*Department of Computer Science*
*University of Rochester*
Rochester, USA
kjain7@u.rochester.edu

Joann Sum
*Department of Computer Science*
*California State University, Fullerton*
Fullerton, USA
Josum@csu.fullerton.edu

Pranav Kapoor
*Department of Computer Science*
*Acadia University*
Wolfville, CA
157998k@acadiau.ca

Dr. Amir Eaman
*Department of Computer Science*
*Acadia University*
Wolfville, CA
amir.eaman@acadiau.ca

*Abstract*—Analysis of Security-Enhanced Linux (SELinux) policies requires extensive manual effort to identify violations and security misconfigurations. Current tools employ mathematical abstractions that, while theoretically sound, produce outputs that practitioners struggle to interpret effectively. Automated approaches using machine learning have shown promise but fail to capture the complex relationships inherent in SELinux policies. Here we present a novel approach combining graph-based policy representation with neural networks to automate SELinux policy analysis. Our method transforms policies into graph structures where nodes represent security contexts and edges capture access relationships, then applies Node2vec to learn continuous feature representations that preserve policy neighborhoods and violation patterns. We develop a flexible policy analysis framework that processes these representations through Random Forest, Support Vector Machine (SVM), and Multi-Layer Perceptron (MLP) models to detect violations. Our experimental results demonstrate that this approach achieves 95% accuracy in identifying security violations while maintaining balanced precision and recall metrics, significantly outperforming existing analysis techniques. Through extensive evaluation on synthetic policy datasets derived from production systems, we show that our method effectively captures diverse violation patterns including separation of duty violations, domain transition issues, and unauthorized access paths. Together, our work presents an efficient approach for automated, interpretable SELinux policy analysis that bridges the gap between theoretical security models and practical policy management.

I. INTRODUCTION

Security-Enhanced Linux (SELinux) implements mandatory access controls (MAC) to enhance the traditional discretionary access control (DAC) model. While DAC bases access decisions solely on user ownership, SELinux requires all subjects (processes) and objects (files, sockets) to satisfy additional policy rules based on their security contexts [1].

Despite SELinux's robust security model, its policy-language complexity creates significant challenges for policy analysis and management. The core problem lies in the disconnect between SELinux's intricate policy framework and the tools available to analyze them. Additionally, the fine-grained nature of SELinux access control necessitates numerous rules, often resulting in policies with thousands of statements that are time-consuming to create and risky to modify [4].

Current SELinux policy analysis tools span a range of approaches. Mathematical proof-based analysis tools often introduce additional layers of abstraction, translating policies first into mathematical logic and then into mathematical models before generating results [3]. This multi-step process, while theoretically sound, produces outputs that are difficult for practitioners to interpret and apply effectively. The SETools suite [6] forms the foundation of basic analysis, providing several key utilities: apol[] offers a graphical interface for exploring and analyzing SELinux policies, allowing users to examine policy components like types, classes, and rules; seinfo[] provides command-line access to statistics and summaries of policy components; and SEsearch enables detailed searching of policy rules with flexible criteria [3]. For policy management, tools like audit2allow, which automatically generates policy rules from denied operations in audit logs, and Semanage, which facilitates the creation and modification of SELinux policy modules without requiring detailed policy language knowledge, enable iterative policy refinement based on system behavior and modular policy administration [3].

Previous efforts, such as Efremov and Shchepetkov's work on runtime verification [12], underscore the need for tools that can map high-level security goals onto lower-level system operations. Likewise, SPLinux [13] demonstrated the value of enforcing information flow policies, but its approach, like others, remains challenging to deploy at scale due to the granularity of policy specifications. Formal models like SELAC [14] provide theoretical frameworks but are not practical for widespread use. This research addresses these gaps by offering a graph-based technique that simplifies the analysis process while maintaining precision.

Recent advancements include formal verification methods using Satisfiability Modulo Theories (SMT) [7], which aim to automatically detect inconsistencies and policy violations. The emergence of machine learning techniques, particularly graph-based approaches using algorithms like node2vec [8], shows promise in identifying anomalous patterns in complex policies. This evolution in analysis methods has led to diverse approaches for policy verification and optimization. SPRT [15] uses prototype networks to classify vulnerabilities and adjust SELinux policies based on vulnerability descriptions, while our research leverages emerging graph-based techniques by employing Neo4j [10] to model policy structures and applying machine learning algorithms to the resulting graph data. Where SPRT [15] focuses on categorizing vulnerabilities to guide policy modifications, our work emphasizes detecting anomalous patterns in policy relationships through graph-based representations and anomaly detection models. These modern approaches attempt to bridge the gap between theoretical rigor and practical usability, though challenges remain in making these solutions accessible to system administrators.

This research aims to develop an automated approach to SELinux policy analysis that is both comprehensive and accessible to security administrators. Our objectives are to: 1) evaluate whether machine learning techniques can effectively automate SELinux policy analysis, and 2) compare the effectiveness of different anomaly detection models in identifying policy violations and misconfigurations.

## II. SELINUX BACKGROUND

SELinux represents a significant advancement in operating system security. This section introduces SELinux's core architecture, explains its security goals and common policy violations, and reviews existing analysis approaches.

### A. SELinux Policy Architecture and Type Enforcement

At the core of SELinux's security model is Type Enforcement (TE), which serves as the primary mechanism for implementing mandatory access controls. Every subject (process) and object (file, socket, etc.) receives a security context label containing user, role, type, and optionally, a level for Multi-Level Security (MLS) implementations [5]. Among these attributes, the type is most crucial for access control decisions, with subjects (typically processes) assigned domain types and objects given resource types. So, our research focuses on SELinux Policy Type Enforcement.

```
allow SourceDType TargetType : class1 {perm1 perm2};
```

Listing 1: Basic Syntax of a SELinux security policy

The rule in Listing 1 represents the fundamental building block of SELinux policy. This allow rule syntax permits the process with domain SourceDType to have actions perm1 or perm2 on the object of type TargetType and object class of class1. An object class specifies the type of resource (such as files, sockets, and directories).

Beyond simple allow rules, the policy language includes type definitions, attributes for grouping related types, and macros for reusable policy blocks [3].

SELinux operates on the principle of least privilege, denying all interactions between types by default unless explicitly permitted through allow rules. A key feature is domain transitions, where processes can securely change their security context when executing certain programs. Consider Listing 2:

```
type_transition httpd_t httpd_exec_t:process httpd_child_t;
```

Listing 2: Example of domain transition rule in SELinux

The rule in Listing 2 indicates that when a process of type 'httpd_t' executes a file labeled 'httpd_exec_t', it transitions to type 'httpd_child_t'. Such transitions enable fine-grained control over process privileges as they execute different programs, ensuring each process operates with the minimum necessary permissions for its current task.

At the application layer, SELinux's type enforcement can be integrated through type-aware interfaces. Applications can use type labels to categorize data and resources, enforcing information flow controls that complement system-level MAC policies. This integration allows applications to enforce their own security rules based on data types while maintaining compatibility with system-wide policies [5].

### B. SELinux Security Goals and Policy Violations

SELinux policies are designed to enforce specific security goals through constraints and access rules. However, policy misconfigurations can lead to violations of these security goals. Understanding common violation types is crucial for maintaining system security.

Separation of Duty (SoD) represents a fundamental security goal where critical operations should be divided among multiple entities. Consider the misconfigured policy in Listing 3:

```
allow financial_data_t audit_log_t:file { read write };
allow audit_log_t financial_data_t:file { write };
```

Listing 3: Example of SoD violation in SELinux policy

This configuration violates SoD by allowing a single process type (financial_process_t) to both modify financial data and write audit logs, potentially enabling fraud through manipulation of both transaction and audit records. A secure configuration would separate these duties as shown in Listing 4:

```
allow financial_process_t financial_data_t:file { read write };
allow audit_process_t audit_log_t:file { write };
```

Listing 4: Example of SoD enforcement in SELinux Policy

Another violation we focus on is domain transition issues. Domain transition issues arise from incomplete or incorrect transition rules between security contexts. For proper operation, domain transitions require specific combinations of entrypoint access, execute permissions, and transition rights. Missing or misconfigured transition rules can prevent legitimate operations or create security vulnerabilities. Policy inconsistencies also manifest through contradictory rules, where conflicting permissions create unpredictable behavior, and through incorrect type usage where security contexts are inappropriately assigned to resources.

Network-related violations, particularly unauthorized network access, represent a distinct threat category where processes may gain unintended network capabilities. Additionally, mislabeled files and processes can lead to both security vulnerabilities and system functionality issues, while missing necessary file access rules for system processes can disrupt essential operations. These violations often interact in complex ways - for example, a combination of mislabeled files and improper privilege assignment could create unauthorized access paths that are difficult to detect through manual inspection [3].

The complexity of SELinux policies means violations can manifest in various ways, from simple permission misconfigurations to subtle interactions between multiple rules. These violations often require sophisticated analysis techniques for detection, as manual inspection becomes impractical with the thousands of rules present in typical SELinux deployments [3]. A comprehensive categorization and analysis of specific violation classes and their detection through our machine learning approach is presented in Section 4.1.

## III. GRAPH-BASED ANALYSIS

Graph-based analysis has emerged as a powerful approach for analyzing complex security policies and access control systems. Recent work by Wu et al. demonstrates its effectiveness in analyzing large-scale security policies [2], while research in cloud computing security has shown graphs to be particularly effective at representing and analyzing complex permission relationships [7]. This methodology has gained traction in security policy analysis due to its ability to represent and process complex relationships efficiently, making it particularly well-suited for analyzing SELinux's intricate policy structures. The effectiveness of graph-based approaches has been further validated by recent studies applying graph neural networks to security policy analysis [9].

Fundamentally, SELinux policies lend themselves naturally to graph representations. In this graph model, types—the core elements of SELinux's Type Enforcement mechanism—can be conceptualized as nodes in a graph. The allow rules that define permitted interactions between these types form the edges connecting these nodes. This mapping provides an intuitive and mathematically rigorous foundation for policy analysis [2].

### A. Benefits of Using Graph For Analysis

One of the primary strengths of graph-based analysis lies in its focus on relationships. SELinux policies are, at their core, about defining and constraining relationships between different entities in a system. Graph structures excel at capturing and representing these relationships, allowing for efficient analysis of access paths and potential information flows. This relational focus aligns closely with the fundamental security questions that policy analysts need to address, such as determining what resources a given process type can access or identifying all potential paths between two types [9,11].

The visual nature of graphs provides another significant advantage. Complex policies that might be difficult to comprehend when expressed as long lists of rules can become much more accessible when visualized as graphs. This visual intuition can help administrators quickly identify patterns, anomalies, or potential security issues that might not be apparent from textual representations alone. Visualization tools based on graph representations have shown promise in enhancing policy comprehension and analysis efficiency [2].

From a computational perspective, graph databases and algorithms offer efficient mechanisms for querying and analyzing complex relationship structures. Traditional relational databases can struggle with the types of recursive queries often needed in security policy analysis, such as finding all possible paths between two types. Graph databases, in contrast, are optimized for such traversals, allowing for more efficient and scalable analysis of large policies [9].

### B. Graph Model for SELinux Policies

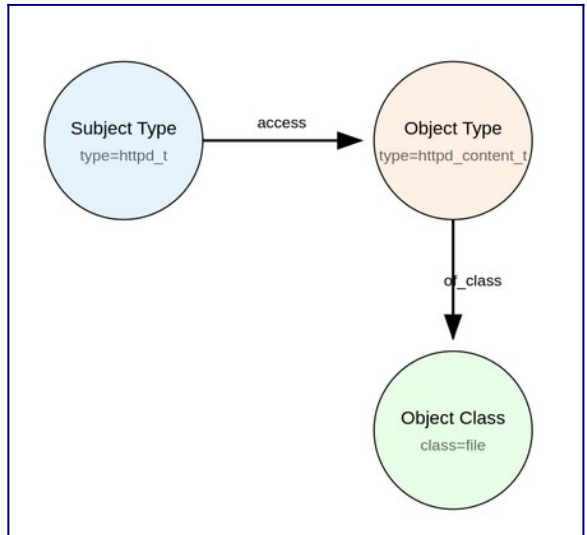

Fig 1: Example of SELinux Type Enforcement Graph Model

Our graph representation of SELinux policies in Figure 1 builds on the model proposed by Eaman et al. [9], which defines three fundamental node types to capture policy relationships. The model represents processes and domains as Subject nodes containing name and type properties, resources as Object nodes with name, type, and class properties, and permission categories as Class nodes storing name and associated permissions. This structure effectively captures the hierarchical nature of SELinux policies, where subjects (processes) interact with objects (resources) according to defined class permissions. The relationships between these nodes directly represent the allow rules in the SELinux policy, with edges from Subject to Object nodes indicating permitted operations under specific class constraints [5]. This graph structure provides a natural representation of SELinux's Type Enforcement mechanism, enabling efficient analysis of permission relationships and policy patterns.

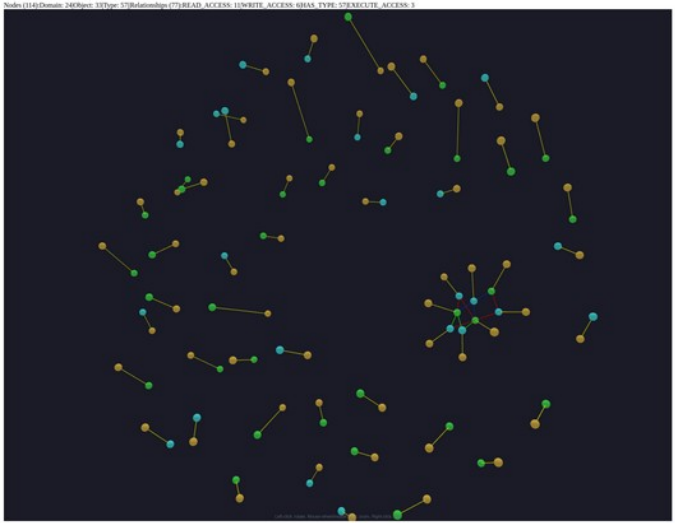

Fig. 2: Graph of Synthetic Policies using Neo4j and Three.js

Expanding on the model in Figure 1, allows for a comprehensive representation of SELinux policies within a graph database, such as in Figure 2. Subject nodes represent processes or domains that can initiate actions. Object nodes represent resources that can be acted upon, such as files or sockets. Class nodes define the types of objects and their associated permissions.

The relationships between these nodes represent the allow rules in the SELinux policy. For instance, an edge from a Subject node to an Object node would indicate that the subject has certain permissions on that object, as defined by the policy [5]. While this graph representation provides an intuitive way to visualize policy relationships, analyzing complex policy patterns requires transforming these graph structures into a format suitable for machine learning algorithms. This transformation process involves several steps: first, converting the policy rules into a graph structure that captures all relevant security relationships; then, encoding this graph structure into numerical features that preserve both local and global policy patterns; and finally, preparing these features for input into machine learning models.

## IV. MACHINE MODEL DEVELOPMENT AND TRAINING

During our initial analysis, we observed that traditional query generation often resulted in very specific queries, which may not encompass all potential errors or violations. In contrast, our violation detection model offers more opportunities to detect new and potentially unforeseen violations. The model we developed is intended to read the policy data and populate a "violation class" column, categorizing each policy rule into one of four violation classes.

Our approach combines graph-based structural analysis with machine learning techniques to automate SELinux policy violation detection. This hybrid approach begins with developing scripts to extract, parse, and import policy data into a graph database, creating the foundation for our machine learning pipeline [3]. We leverage Node2vec, a deep learning algorithm that generates continuous feature representations for nodes in networks [8], to transform our graph structures into vector embeddings that capture both structural and semantic policy relationships. This vectorization step is crucial for enabling our subsequent machine learning analysis.

The training data for our models is constructed from three distinct policy aspects: SELinux Transition Graphs capturing entity interactions and transitions, attribute graphs representing entity relationships, and Object Class Graphs encoding permission hierarchies. These graph representations are then processed through our machine learning pipeline, employing Random Forest, Support Vector Machine (SVM), and Multi-Layer Perceptron (MLP) Neural Network models to detect policy violations. This combination of graph-based representation and modern machine learning techniques provides a robust framework for automated policy analysis, capable of handling the scale and complexity of real-world SELinux deployments [3], [8].

### A. Defining Violation Classes

Security administrators can apply security goals through policies. These goals are expressed through security constraints which, along with access rules, specify access decisions, i.e., to grant or deny an access request.

Common policy violations manifest in several ways in SELinux systems. For example, there are Separation of Duty Violations as outlined previously in Section 2.2. Additionally, there are other violation classes we classify while training our model such as contradictions, which arise when conflicting rules create unpredictable behavior, such as when one rule allows access while another denies it. Also, missing rule violations, exemplified by cases where the absence of network access restrictions creates security gaps, can leave systems vulnerable [3]. And incorrect Type Usage violations occur when inappropriate types are assigned to resources, such as labeling system binaries with user data types. Another important concept is overly permissive rules which create unnecessary attack surfaces by granting excessive permissions beyond operational requirements. Domain transition issues arise when any of three required conditions fail: entrypoint access to exec file type, execute access to entry point file type, and transition access to new domain type. Finally, Mislabeled Files and Processes, where incorrect context assignments lead to unintended access restrictions or permissions, can severely impact system functionality [3].

By carefully analyzing for these potential anomalies, administrators can identify and rectify policy misconfigurations, ensuring that SELinux policies are correctly implemented and aligned with the intended security goals. In our models we define violation classes to address these common policy violations, finetuning them to be more specific.

0: No anomalies

1: Separation of Duty (SoD) violation - single subject with read and write access to sensitive data

2: Improper privilege assignment

3: Critical system file modification

4: Incorrect type usage

5: Domain transition issues

6: Mislabeled files or processes

7: Unauthorized network access

8: Separation of Duty (SoD) violation - single subject with access to multiple mutually exclusive roles

9: Contradictory type transitions for the same process

10: Missing necessary file access for system processes

Listing 5: Specific Violation Classes Used when Training Models

The anomalies we aim to detect can be categorized into ten distinct violation classes, as shown in Listing 5. These range from access control violations to type assignment issues and system integrity concerns. In terms of access control, we detect Separation of Duty (SoD) violations where a single subject has both read and write access to sensitive data (Type 1), or access to multiple mutually exclusive roles (Type 8). We also identify unauthorized network access patterns (Type 7) that could indicate security bypasses. For type assignment issues, our models detect incorrect type usage (Type 4), mislabeled files or processes (Type 6), and contradictory type transitions that create ambiguous process contexts (Type 9). System integrity concerns include improper privilege assignments (Type 2), unauthorized critical system file modification access (Type 3), domain transition issues that could enable privilege escalation (Type 5), and missing necessary file access for system processes (Type 10). Type 0 serves as our baseline case, representing properly configured policies with no detected anomalies. Each violation class corresponds to specific patterns in the policy graph structure that our detection models are trained to identify. These examples were specifically chosen to represent realistic misconfigurations that security administrators might encounter in production environments.

### B. Dataset Construction and Preparation

We constructed a synthetic dataset to enable controlled evaluation of our violation detection models. Initial analysis of Fedora 39 and Ubuntu Server SELinux policies provided templates for policy structure and common patterns. Using these patterns, we generated synthetic examples representing each violation class described in Section 2.2.

The dataset focuses on server security contexts, including web servers and database systems, where policy violations present significant risks. We developed examples covering all ten violation categories from Listing 5. Each synthetic policy isolates specific security properties to enable precise evaluation of our detection capabilities. The examples follow the syntax shown in Listing 1 while incorporating violations like the SoD issues demonstrated in Listings 3 and 4.

Our validation process employed graph-based modeling to verify that each synthetic example correctly represented its intended security properties. This approach allowed us to confirm that our synthetic policies exhibited the structural characteristics of real SELinux policies while containing well-defined violations suitable for training our machine learning models.

This synthetic approach, while not utilizing complete production policies, provided several advantages for our research. It allowed us to create a balanced dataset with well-understood properties, avoided potential security concerns associated with using production policies, and enabled us to systematically evaluate our models' detection capabilities across different violation types. The controlled nature of the synthetic data also facilitated more precise evaluation of our models' performance characteristics and generalization capabilities.

### C. Results

Employing our dataset preparation method from Section 4.2, we then selected and evaluated three distinct machine learning models for policy violation detection: Random Forest, SVM, and MLP models. Random Forest was chosen for its ability to handle high-dimensional data and capture complex rule interactions through ensemble learning of decision trees, achieving 93% accuracy with balanced precision and recall (0.93/0.93/0.93) in violation detection. SVM was selected for its effectiveness in handling binary and multi-class classification problems with clear decision boundaries, demonstrating 92% accuracy with strong performance metrics (0.93/0.92/0.92). The MLP Neural Network was included for its capacity to learn complex non-linear relationships in the policy data, ultimately providing the best performance with 95% accuracy and highest precision/recall scores (0.95/0.97/0.95).

For model training, when tested on our largest dataset of 455 policy rules, Node2vec-based models achieved consistently higher performance, with the MLP model reaching 95% accuracy. Node2vec's superior performance stems from its ability to preserve both local and global graph structures through its flexible random walk strategy, which proved crucial for capturing the complex relationships in SELinux policies [8]. This approach generates rich feature representations that encode both structural and semantic aspects of policy relationships, enabling our models to better identify policy violations across varying contexts and scales.

To evaluate the effectiveness of our approach, we conducted a series of experiments with increasing complexity. Starting with a basic set of violation classes, we progressively refined our classification schema based on model performance and real-world policy patterns. This iterative process helped us understand both the capabilities and limitations of different model architectures while working with SELinux policy data.

We initially tested our models on 5 violation classes and with a smaller dataset of 125 policy rules, including the control class '0' representing policies without a violation. The model we developed is intended to read the data and populate the violation class column.

Here are some key results for a dataset of 125 policy rules:

Table 1: Results for 125 Policy Rules

| Model | Accuracy | Macro Avg (P/R/F1) | Weighted Avg (P/R/F1) |
|---|---|---|---|
| Random Forest | 0.79 | 0.85/0.77/0.79 | 0.83/0.79/0.79 |
| SVM | 0.68 | 0.80/0.65/0.63 | 0.78/0.68/0.65 |
| MLP | 0.86 | 0.89/0.85/0.86 | 0.88/0.86/0.85 |

Our initial approach with 5 violation classes proved too broad, leading to high false positives particularly in classes like "Separation of Duty Violations" and "Domain Transition Issues". For example, the general SoD violation class encompassed both file access violations and role-based violations, which exhibited different structural patterns in the policy graphs. This led us to split SoD into two distinct classes: "SoD violation - single subject with read and write access to sensitive data" and "SoD violation - single subject with access to multiple mutually exclusive roles".

So, we then split up the violation classes into more specific examples with patterns we believed the models would pick up more accurately. Our next test used 10 violation classes: No anomalies, Separation of Duty (SoD) violation, Overly permissive access, Improper privilege assignment, Critical system file modification, Contradictory rules, Missing necessary rules, Incorrect type usage, Domain transition issues, Unauthorized network access. And we expanded the dataset to include 401 different types of rules with a balanced amount of each violatioin class.

Table 2: Results for Initial 10 Violation Classes

| Model | Accuracy | Macro Avg (P/R/F1) | Weighted Avg (P/R/F1) |
|---|---|---|---|
| Random Forest | 0.85 | 0.80/0.77/0.77 | 0.87/0.85/0.85 |
| SVM | 0.73 | 0.43/0.47/0.45 | 0.68/0.73/0.69 |
| MLP | 0.83 | 0.78/0.74/0.73 | 0.86/0.83/0.83 |
| Stacking Ensemble | 0.83 | 0.77/0.77/0.77 | 0.85/0.83/0.84 |

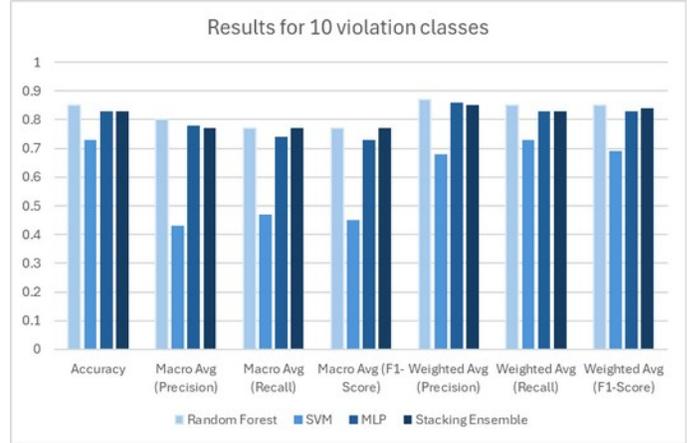

Fig. 3: Bar Graph of Initial 10 Violation Class' Results

With these classes, our Random Forest model achieved an accuracy of 0.85, with a macro average precision/recall/F1 of 0.80/0.77/0.77 and a weighted average of 0.87/0.85/0.85. The SVM model had an accuracy of 0.73, while the MLP achieved 0.83. We also implemented a Stacking Ensemble model, which achieved an accuracy of 0.83.

Due to poor results from classes related to Separation of Duty, Contradictory Rules, and Missing Rules, we decided to expand our classification to 16 more specific violation classes. This included separating the problematic classes into more detailed categories. And the dataset included 455 different example policies.

Table 3: Results for 16 Violation Classes

| Model | Accuracy | Macro Avg (P/R/F1) | Weighted Avg (P/R/F1) |
|---|---|---|---|
| Random Forest | 0.87 | 0.85/0.83/0.82 | 0.88/0.87/0.86 |
| SVM | 0.82 | 0.84/0.78/0.77 | 0.85/0.82/0.80 |
| MLP | 0.86 | 0.84/0.82/0.82 | 0.87/0.86/0.85 |

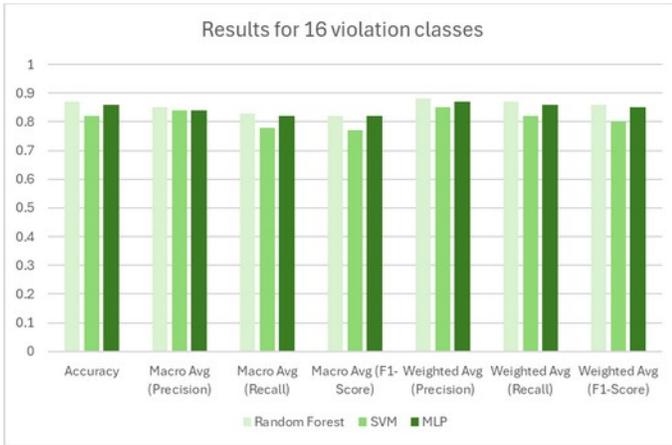

Fig. 4: Bar Graph of 16 Violation Class' Results

With this refined classification, our model performance improved. The Random Forest model now achieved an accuracy of 0.87, with a macro average precision/recall/F1 of 0.85/0.83/0.82 and a weighted average of 0.88/0.87/0.86. The SVM model improved to an accuracy of 0.82, and the MLP reached 0.86.

These results demonstrate the importance of specific and well-defined violation classes in improving the accuracy of our models. Initial expansion to 16 classes allowed us to distinguish between subtle variations in policy violations. For example, we split contradictory rules into two distinct classes: "Contradictory allow and deny rules for same subject-object-permission combination" and "Contradictory type transitions for the same process". This separation helped identify specific patterns in policy misconfigurations. However, we found that this fine-grained separation sometimes caused the models to miss broader patterns of contradiction. While the models could identify exact matches of contradictory rules, they often failed to detect conceptually similar contradictions that differed slightly in structure. This observation led us to first consolidate these into a single "Contradictory rules" class in our 15-class model, while maintaining the same dataset, improving the model's ability to detect contradictions in various forms. Additionally, all models struggle with class 14 (Missing necessary file access for system processes), showing particularly low recall as shown in Table 6:

Table 4: Recall Score for violation class 14 (Missing necessary file access for system processes)

| Model | Recall Score |
| --- | --- |
| Random Forest | 0.25 |
| SVM | 0.06 |
| MLP | 0.25 |

Despite achieving higher overall accuracies in the datasets using 16 violation classes, as shown in Table 4 and 5, the models exhibited difficulty in detecting certain types of violations, particularly those that are rare or complex, such as "Missing necessary file access for system processes." The models' performance on these classes was notably lower, indicating a need for further refinement.

Our refinement from 16 to 10 classes involved strategic consolidation of related violation types. The overly permissive access class (originally class 2) was merged into improper privilege assignment, as our models showed significant overlap in detecting these patterns. Similarly, we consolidated all contradictory rule violations (original classes 5, 12, and 13) into a single "Contradictory type transitions" class, as these violations shared common structural patterns in the policy graphs.

A significant consolidation occurred with system access violations. The original separate classes for missing port access (class 6), file access (class 14), directory access (class 15), and network access (class 16) were combined into a single "Missing necessary file access" class. This consolidation was driven by our observation that the models struggled with overly specific access violations, showing particularly low recall scores for these classes (around 0.25 for both Random Forest and MLP).

Table 5: Results for 10 Refined Violation Classes

| Model | Accuracy | Macro Avg (P/R/F1) | Weighted Avg (P/R/F1) |
| --- | --- | --- | --- |
| Random Forest | 0.93 | 0.93/0.93/0.93 | 0.94/0.93/0.93 |
| SVM | 0.92 | 0.93/0.92/0.92 | 0.92/0.92/0.92 |
| MLP | 0.95 | 0.95/0.97/0.95 | 0.96/0.95/0.95 |

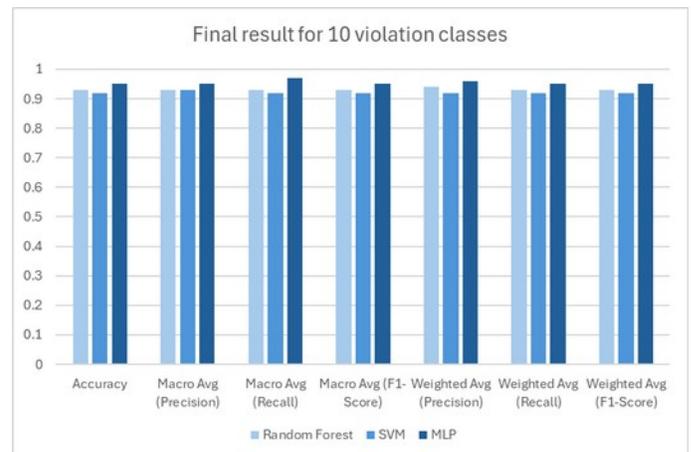

Fig. 5: Bar Graph of Final Refined 10 Violation Class' Results

The consolidated 10 classes in Figure 5 showed improved detection rates while maintaining the ability to identify critical access violations. By combining related violations into broader but still meaningful categories, we achieved better generalization while maintaining the ability to detect specific types of policy misconfigurations. The final 10-class model showed more balanced performance across all categories, with improved precision and recall metrics compared to the more granular 16-class model. Thus, we went with refined 10 classes as our final model.

According to Table 5, all three models perform very well, with accuracies ranging from 0.92 to 0.95. The MLP model achieves the highest accuracy (0.95), followed by Random Forest (0.93) and SVM (0.92). The MLP model shows strong performance across all metrics, with macro-averaged precision, recall, and F1-scores of 0.95, 0.97, and 0.95 respectively, and weighted averages of 0.96, 0.95, and 0.95. The Random Forest model demonstrates consistent performance with both macro and weighted averages around 0.93, while the SVM model shows similar consistency with metrics around 0.92. This indicates that all models maintain good balance across different violation classes, though the MLP model appears to have a slight edge in overall performance.

## V. Conclusion and Future Direction

The findings from our research underscore the efficacy of graph-based machine learning models in automating the analysis of SELinux policies. By representing policies as graphs and applying models like Random Forests, SVM, and MLP Neural Networks, we achieved high accuracy in detecting policy violations. Notably, the MLP model consistently demonstrated robust performance across varying dataset sizes and classification schemes, achieving accuracies up to 96%.

While our research demonstrates the effectiveness of machine learning models in detecting SELinux policy violations, further work is needed to evaluate real-world performance implications. System administrators could potentially integrate this approach into their workflows through automated policy analysis, but actual deployment would require careful performance testing and optimization. The computational overhead of graph construction, embedding generation, and model inference in production environments with large policy sets remains to be thoroughly benchmarked. Future research should focus on measuring these performance characteristics across different scales of deployment, from small systems to enterprise environments with hundreds of thousands of rules. This performance analysis would help determine whether the approach is better suited for periodic policy audits or if it could be feasibly implemented as part of real-time policy validation.

Additionally, reinforcement learning could enable more dynamic policy analysis, with agents learning to identify violations through interaction with policy environments. Unlike our current supervised learning approach, RL agents could potentially discover novel attack vectors and policy weaknesses by simulating various security scenarios. The integration of machine learning techniques with advanced methodologies presents exciting opportunities. Large Language Models (LLMs) with Retrieval Augmented Generation (RAG) could assist in policy interpretation, generate human-readable explanations of violations, and potentially help bridge the gap between high-level security requirements and low-level policy specifications [2].

A particularly promising direction is better integration of type-based enforcement at both system and application levels. This approach could enable applications to enforce fine-grained security decisions internally based on type labels while maintaining compatibility with system-wide policies. By allowing applications to dynamically interact with SELinux's security contexts, this model could provide more flexible and granular security enforcement [5].

Our results suggest that machine learning-based policy analysis can effectively bridge the gap between SELinux's powerful security features and administrators' practical needs. The combination of graph-based representation, sophisticated machine learning models, and automated analysis tools provides a promising framework for enhancing SELinux policy management. While challenges remain in areas such as model interpretability and computational resources, our approach demonstrates significant potential for improving security policy analysis in complex Linux environments. The ultimate goal is to develop a comprehensive, automated system that makes SELinux policy analysis more accessible while maintaining the security guarantees that make SELinux valuable for system security.

## Acknowledgment

We express our gratitude to Prof. Daniel Jackson (MIT CSAIL) for his valuable insights on applying formal verification approaches, particularly his suggestions around using Alloy for policy verification and property checking. His guidance helped shape our methodology for ensuring policy correctness. We also thank the engineers at Google and Red Hat for their constructive feedback and look forward to future collaborations as we continue developing this work.